\documentclass{article}
\usepackage{arxiv}
\usepackage{amsmath}
\usepackage[utf8]{inputenc} 
\usepackage[T1]{fontenc}    
\usepackage{hyperref}       
\usepackage{url}            
\usepackage{booktabs}       
\usepackage{amsfonts}       
\usepackage{nicefrac}       
\usepackage{microtype}      
\usepackage{lipsum}
\usepackage{graphicx}
\usepackage{xcolor}
\usepackage{subcaption}
\usepackage{makecell}
\usepackage{listings}
\usepackage{pdfpages}
\usepackage{multibib}

\definecolor{codegreen}{rgb}{0,0.6,0}
\definecolor{codegray}{rgb}{0.5,0.5,0.5}
\definecolor{codepurple}{rgb}{0.58,0,0.82}
\definecolor{backcolour}{rgb}{0.95,0.95,0.92}
\lstdefinestyle{mystyle}{commentstyle=\color{codegreen},
  keywordstyle=\color{magenta},
  numberstyle=\tiny\color{codegray},
  stringstyle=\color{codepurple},
  basicstyle=\ttfamily\footnotesize,
  breakatwhitespace=false,         
  breaklines=true,                 
  captionpos=b,                    
  keepspaces=true,                 
  numbers=left,                    
  numbersep=5pt,                  
  showspaces=false,                
  showstringspaces=false,
  showtabs=false,                  
  tabsize=2
}

\title{Search for temporal cell segmentation robustness in phase-contrast microscopy videos}

\author{
    Estibaliz~Gómez-de-Mariscal \\
    Bioengineering and Aerospace,\\
    Engineering Department,\\
    Universidad Carlos III de Madrid and \\
    Instituto de Investigación Sanitaria \\
    Gregorio Marañón, Madrid, Spain\\
\And
    Hasini Jayatilaka \\
    AtlasXomics Inc.,\\
    New Haven, USA
\And
    \"{O}zg\"{u}n Çiçek\\
    Department of Computer Science,\\ 
    Albert-Ludwigs-University, \\
    Freiburg, Germany
\And
    Thomas Brox\\
    Department of Computer Science,\\ 
    Albert-Ludwigs-University, \\
    Freiburg, Germany
\And
    Denis Wirtz\\
    Department of Chemical and\\
    Biomolecular Engineering,\\
    Institute for Nanobiotechnology,\\
    The Johns Hopkins University and\\
    Department of Oncology,\\
    The Johns Hopkins University \\
    School of Medicine,\\ Baltimore, USA
\And
    Arrate~Muñoz-Barrutia \\
    Bioengineering and Aerospace,\\
    Engineering Department,\\
    Universidad Carlos III de Madrid and \\
    Instituto de Investigación Sanitaria \\
    Gregorio Marañón, Madrid, Spain\\
    \texttt{mamunozb@ing.uc3m.es} \\
}

\begin{document}
\maketitle

\begin{abstract}
\textbf{Studying cell morphology changes in time is critical to understanding cell migration mechanisms. In this work, we present a deep learning-based workflow to segment cancer cells embedded in $3$D collagen matrices and imaged with phase-contrast microscopy. Our approach uses transfer learning and recurrent convolutional long-short term memory units to exploit the temporal information from the past and provide a consistent segmentation result. 
Lastly, we propose a geometrical-characterization approach to studying cancer cell morphology. Our approach provides stable results in time, and it is robust to the different weight initialization or training data sampling. We introduce a new annotated dataset for $2$D cell segmentation and tracking, and an open-source implementation to replicate the experiments or adapt them to new image processing problems.} 
\end{abstract}


\section{Introduction}

Metastasis, the main cause of death caused by cancer, refers to the process in which cells spread from the primary tumor to adjacent tissues, proliferate and invade healthy organs. Thus, understanding the mechanisms driving cancer cell migration is critical to characterize highly metastatic cells, develop new efficient treatments and improve precision medicine. Due to the experimental complexity, most quantitative cell migration studies are performed on $2$D cell culture experiments, while cell migration naturally occurs in $3$D environments (i.e., the extracellular matrix (ECM)). Instead of the lamellipodia or filopodia usually observed in $2$D substrates, cells migrating in $3$D substrates (\textit{e.g.}, collagen type I matrices) form dendritic protrusions to anchor, exert forces, and propel. Consequently, there is a growing interest to study cell morphology and motility in $3$D environments~\cite{doyle2015local, Wu2018, Jayatilaka2018, doyle20221}.

Sample phototoxicity and photobleaching are well-known constrains in fluorescence microscopy. Thus, prolonged time-lapse videos are customarily acquired using brightfield microscopy techniques such as phase contrast, at the expense of low contrast between the cell and the background. To mimic the mechanical properties of the ECM, cells are commonly embedded in matrices of collagen type I~\cite{Wu2018, doyle2015local}. Additionally, the brightfield images of collagen matrices are full of artifacts and quite heterogeneous due to the polymerization of the collagen fibers resulting in images full of artifacts. See Figure~\ref{fig: labexperiment differences} in the Appendix. The common setup to study $3$D cell motility consists of acquiring images of a fixed focal plane placed in the middle of the collagen matrix where the cells are embedded. Therefore, the cell movement is assessed in the cited $xy$-plane under the premise that $3$D cell motility is isotropic~\cite{Wu2018}. Note that while the images are $2$D, cells are migrating in $3$D, so they can exit and enter the plane of focus, fixed along with the videos. 

The manual annotation of cells in such long phase-contrast microscopy time-lapse videos is tedious and non-viable. Hence, implementing a robust computational tool for the automatic quantification of cell morpho-dynamics is critical to deciphering the mechanisms that drive cancer cell migration in metastasis.

Deep learning (DL) approaches are considered the state-of-the-art for processing images with large inter-variability and intra-variability. Usiigaci \textit{et al.}~\cite{tsai2019usiigaci} released one of the first methods (based on a Mask R-CNN~\cite{he2017mask}) that achieve accurate cell instance segmentation in phase-contrast microscopy images. Filip \textit{et al.}~\cite{lux2020cell} combine DL segmentation and cell detection for a posterior watershed instance segmentation. Pixel embedding approaches~\cite{payer2018instance, lalit2021embedding} are efficient instance segmentation strategies, especially for highly packed cells. Other works show accurate results for binary segmentation using recurrences such as the convolutional Long-Short Term Memory (LSTM) U-Net~\cite{arbelle2019microscopy} or the recurrent U-Net~\cite{wang2019recurrent}. Additionally, DeepCellKiosk~\cite{bannon2021deepcell} is a cloud-based toolbox to train and deploy DL models for cell segmentation, detection and tracking, and to ease the use of large datasets. Some of the previously mentioned methods (\emph{e.g.}, Mask R-CNN, convolutional LSTM U-Net, cosine-embedding) have high-computational requirements when adapting their implementations to new experiments or data types. Others are not optimized for a low cell/background ratio (\emph{e.g.} DeepCellKiosk models for instance segmentation, the approach of Filip \textbf{\emph{et al.}}). Additionally, most previous contributions focus on defining the convolutional neural network (CNN) architecture and, sometimes, the loss function rather than the training strategy. Aspects to consider for the latter are the use of pre-trained encoders, dealing with data imbalance, or avoiding the creation of artifacts during data augmentation. From our in-house experiments, we realized that the quality of the ground truth and the training strategy was the most important factors to get accurate and robust results.

To analyze the relationship between cell morphology and motility, we need to detect cell protrusions. There exist several works in the literature addressing a closely related problem with cell filopodium~\cite{mavska2013automatic, Tsygankov2014, barry2015, filopodyan2017, Castilla2019, Bagonis2019}. The previously cited methods work under the hypothesis that a well-defined frontier between the cell and the filopodia exists. However, setting the limit between the cell body and the cell protrusions is still an open question. This inconvenience can be circumvent by detecting instead the protrusion tips. 

To summarize, in this work, we propose a deep learning (DL)-based bioimage processing workflow: (1) to segment cells on phase-contrast microscopy videos, and (2) to detect their protrusions. We first build a consistent and heterogeneous ground truth image dataset. Then, we propose a workflow that combines deep CNN and geometrical analysis of the cell morphology to quantify cellular protrusions automatically.

%
\begin{figure}[t!]

\centerline{\includegraphics[width=\textwidth]{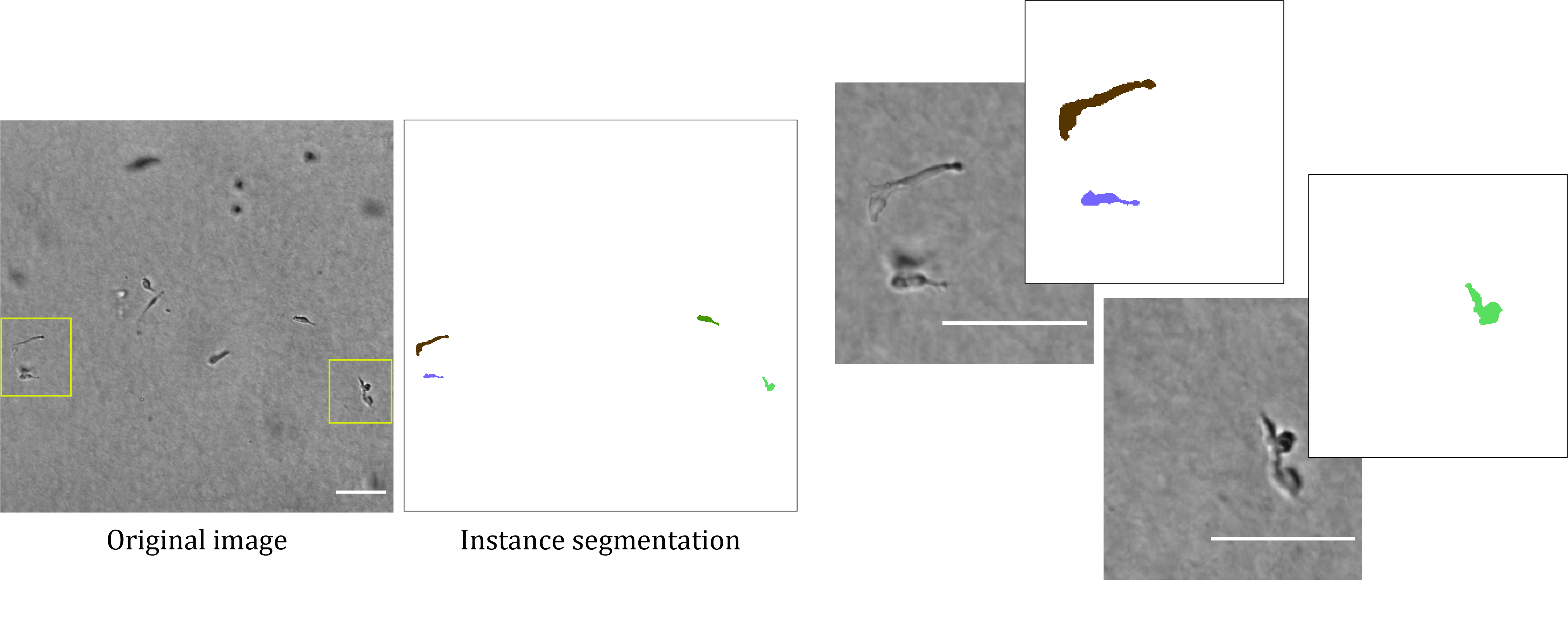}}

\caption[Manual annotations of phase contrast microscopy images of human fibrosarcoma cells embedded in a $3$D collagen Type I matrix.]{
Sample image of a human fibrosarcoma (HT1080WT) cells embedded in a 3D collagen Type I matrix, From left to right: Phase contrast microscopy image; manual annotations of cells in focus in (a); Zoomed versions of the crops shown in the yellow boxes in (a). Cells out of focus are discarded. Scale bars of $100 \mu m$.}
\label{fig: manual annotations}
\end{figure}

\section{Materials}
\label{chap1sect: materials}

We build the ground truth dataset from the phase contrast microscopy videos used in~\cite{Jayatilaka2018}: Human fibrosarcoma HT1080WT (ATCC) cells at low cell densities embedded in $3$D collagen type I matrices. The time-lapse videos were recorded every $2$~minutes for $16.7$~hours and covered a field of view of $ 1002 \:\mbox{pixels} \times 1004 \:\mbox{pixels}$ with a pixel size of $0.802\mu m / pixel$ The videos were pre-processed to correct frame-to-frame drift artifacts, resulting in a final size of $983\:\mbox{pixels} \times 985\:\mbox{pixels}$. All the details are given in the Appendix~A.

The ground truth is built with heterogeneous and independent videos. We choose $27$ videos from independent replicates in which mitosis and apoptosis events, and different cell morphology and migration patterns were present. We ensure that cells touching each other and migrating faster or slower are also present. The variation of the collagen matrix under the microscope is also considered. We subtract short sections between $3$ and $100$ frames from the videos to gather the mentioned events. Finally, our ground truth data consists of $56$ short videos, resulting in a total of $992$ frames. The instance segmentation of focused cells over time are manually annotated and uniquely labeled preserving the tracking information. Three experts annotated the videos and a majority voting method as described in \cite{Ulman2017} was applied to combine the three annotations and build a consensus ground truth. The ground truth data can be accessed in \url{https://zenodo.org/record/5777994}.


\section{Cell segmentation}

\subsection{Methodology}

We propose to train a U-Net-shaped encoder-decoder with a pre-trained encoder, depth-wise separable convolutions, and convolutional long short term memory (LSTM)~\cite{shi2015convolutional} units to obtain a binary mask with the cells being represented by the foreground pixels. 
\begin{figure}[t]
\centerline{\includegraphics[width=\textwidth]{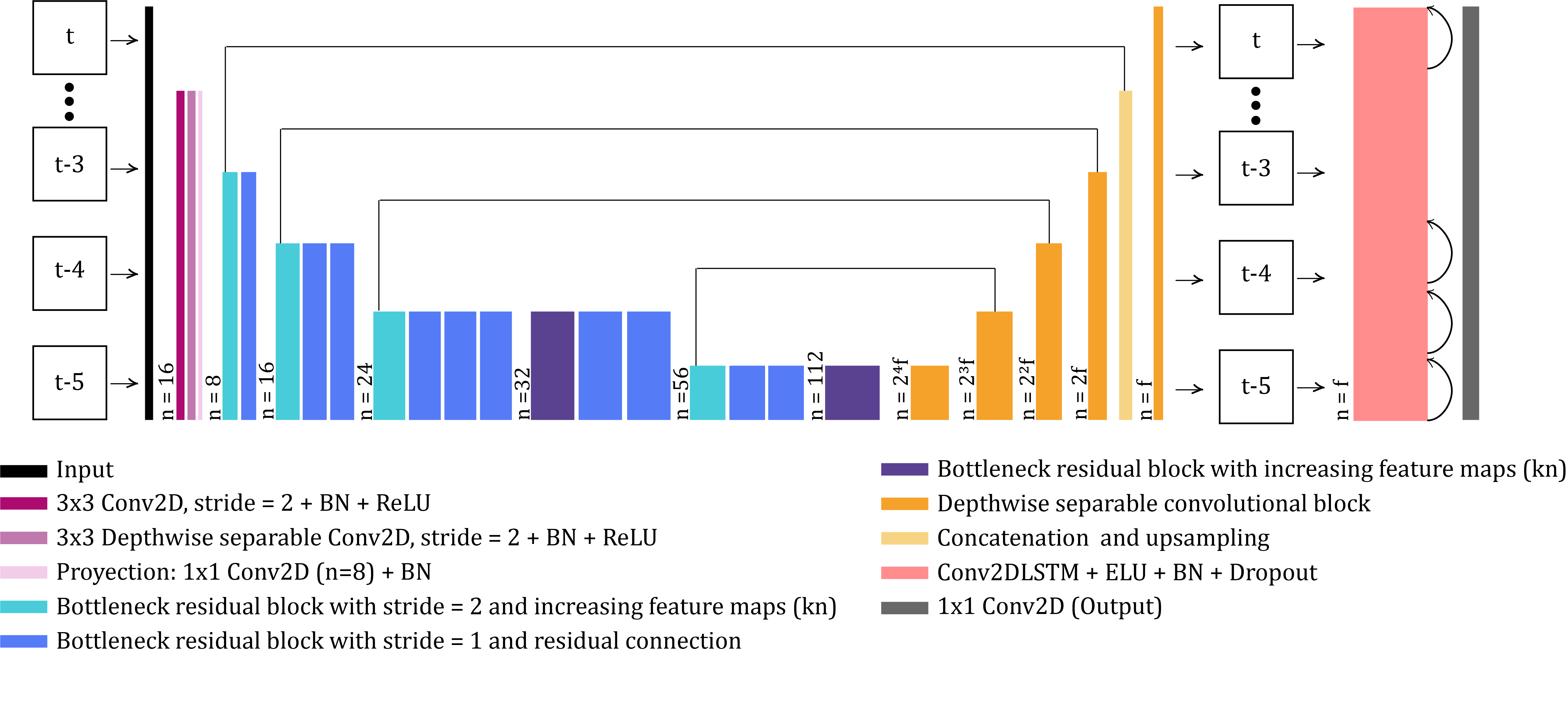}}
\caption[Convolutional neural network (CNN) architecture exploits temporal and spatial information.]{Convolutional neural network (CNN) architecture exploits spatio-temporal information. The architecture processes $N$ consecutive frames ($N=5$) to infer the segmentation of the last one. The encoder-decoder processes the input frames as a batch. The last layer uses a ConvLSTM to process the time-frames recurrently. The encoder is a pre-trained MobileNetV2~\cite{sandler2018mobilenetv2} with $35\%$ of its original width (i.e., number of filters) and skip connections to the corresponding blocks in the decoder. Each block of the decoder is formed by a depth-wise separable convolutional block (see Figure~\ref{fig: CNN blocks} in the Appendix). The number of filters in the decoder and the ConvLSTM depend on the parameter $f$.}
\label{fig: CNN architecture}
\end{figure}

We want to introduce the temporal consistency as we, as manual annotators, needed it to achieve a robust segmentation. We identify two possible approaches: (1) using $3$D convolutions (2D images and the time), and (2) using recurrent attributes in the network. We claim that a system able to discard the noise and artifacts and detect the movement of the cell will be sufficient to determine the cellular shape accurately. Cell movement can be easily detected by the shape differences from frame to frame. Exploiting the $3$D-connectivity of a $3$D U-Net requires a huge amount of memory while the cell movement can be easily detected by the local shape differences between frames. Hence, inspired by the work of Arbelle \emph{et al.}~\cite{arbelle2019microscopy}, we decided to build a semi-recurrent $2$D U-Net (architecture, Figure~\ref{fig: CNN architecture}; convolutional blocks, Figure~\ref{fig: CNN blocks} in the appendix). In~\cite{arbelle2019microscopy}, the authors add a convolutional long short term memory (ConvLSTM) on each level of the encoder, so the temporal information is encoded together with the spatial one. As we use a pre-trained encoder, the ConvLSTM units should go in the decoder layers. We use a single ConvLSTM layer at the end of the encoder-decoder to optimize the memory usage and adapt to the limitations of our hardware training systems. Thus, the number of parameters is not significantly increased despite the recurrent layers. Separable depth-wise convolutions are also introduced to optimize the memory usage, as proposed in~\cite{howard2017mobilenets} (See Figure~\ref{fig: CNN architecture}). The last recurrent layer will sequentially analyze the frames of an input video once the $2$D U-Net processes them. With this approach, we expect to improve the robustness to intra-cell variations along time (i.e., caused by cell exiting and entering the plane of focus) and the network's output to be more consistent at the cell edges where the protrusions appear and disappear. The proposed architecture provides segmentation for the frame at time $t$ based on the information in the frames $(t-k, ..., t)$ for a chosen time window $k$. The architecture is as follows: \textbf{$2$D U-Net shaped encoder-decoder}: The encoder is a MobileNetV2~\cite{sandler2018mobilenetv2} pre-trained on the ImageNet~\cite{deng2009imagenet} dataset for image classification with $35\%$ of the filters of its convolutional layers and with skip connections. A decoder with separable depth-wise convolutions is connected to the skip connections of the MobileNetV2. A \textbf{recurrent Conv2D-LSTM layer} that will process the output frames of the encoder-decoder in the time order. A \textbf{final $2$D convolutional layer} with two feature maps to obtain the pixel classification (background and foreground). 

\textbf{Training data sampling strategy:} A common practice to apply data augmentation (DA), is to crop the original images in a large number of patches and modify these patches with random image transformations. Some DA transformations use image mirroring or zero-padding in the image borders, which adds many unrealistic artifacts that would prevent our CNNs from learning correctly. Frequently, the patches are cropped uniformly along the image regardless of the foreground-background ratio. Our proposal consists of first transforming the image and then cropping a patch using a probability distribution function that deals with the foreground-background ratio. The probability distribution function is commonly used in statistical learning when there is data imbalance. In our case, we define the pixel probability distribution function by setting a weight of $50000$ and $1$ to the pixels in the foreground and background, respectively. Note that the ratio of foreground to background is a maximum of $0.06$ in the best case. After setting the weights, the image is normalized, so the sum of all the pixels equals $1$, which means that we create a probability distribution function for all the pixels in the image. A random pixel drawn using this probability distribution is set as the centroid of the patch that will enter the network during training. With this configuration, we ensure to generate cell-containing patches most of the time, and just in a few cases, background patches. The previous method has been implemented in Python using TensorFlow. The code is freely available at \url{https://github.com/esgomezm/microscopy-dl-suite-tf}.\\

\subsection{Experimental results}

A set of $28$ patches from the raw data in the training set were used as a validation set. Those patches remain the same for all our experiments, while the training data is randomly transformed on each iteration. Each input image intensity is normalized using the percentile normalization with $0.1$ and $99.1$ percentages for the lower and upper bounds. The models are trained in two steps: 
\begin{itemize}
    \item[1.] Transfer-learning to our segmentation model of the MobileNetV2 weights for image classification. In this step, the pre-trained encoder is frozen, the decoder is randomly initialized, and the model is trained for 2$K$ epochs. This model (i.e., the decoder) is trained to learn the segmentation task using a large learning rate $(0.005)$. The proposed transfer-learning approach is not sensible to the decoder initialization and the data sampling strategy (see Appendix). 
    \item[2.] We unfreeze the encoder and fine-tune all the weights using a lower learning rate $(0.0001)$. We train for 2$K$ epochs (see Table~\ref{table: B and BT architectures}, and Figures~\ref{fig: result bt2, bt3 and bt5} and~\ref{fig:modeltraining}). 
\end{itemize}
We evaluate five different configurations of the architecture B$i$, $i=1,...,5$ by changing the depth and sizes of the convolutional layers in the decoder. The main architectural details are given in Table~\ref{table: B and BT architectures}. Notice that the number of trainable parameters prevents us from testing a batch size larger than two using our computational resources (see the Appendix). The decoders of the CNN setups in Table~\ref{table: B and BT architectures} are randomly initialized using the Glorot Uniform initializer~\cite{glorot2010understanding}. We choose exponential linear units (ELUs) as the activation of the convolutional layers in the decoder except for the output layers, which are not activated. We apply the categorical cross-entropy loss function given by Equation \ref{eq: loss cell} directly to the logits of the last layer. The latter is recommended for a more numerically stable computation of the gradients\footnote{\href{https://www.tensorflow.org/api_docs/python/tf/keras/losses/SparseCategoricalCrossentropy}{https://www.tensorflow.org/api\_docs/python/tf/keras/losses/SparseCategoricalCrossentropy}}. ADAM is the optimizer chosen, Equation~\ref{eq: adam}. During the training, we evaluate the binary segmentation with the Jaccard index, Equation~\ref{eq: jaccard}. The accuracy is assessed on the test set. The binary segmentation is evaluated using the Evaluation Software provided in the Cell Tracking Challenge~\cite{Ulman2017} (SEG measure). The Appendix describes the total SEG that accounts for all the videos in the test set.

\begin{figure}[htpb]
    \centering
    \includegraphics[width=1\textwidth]{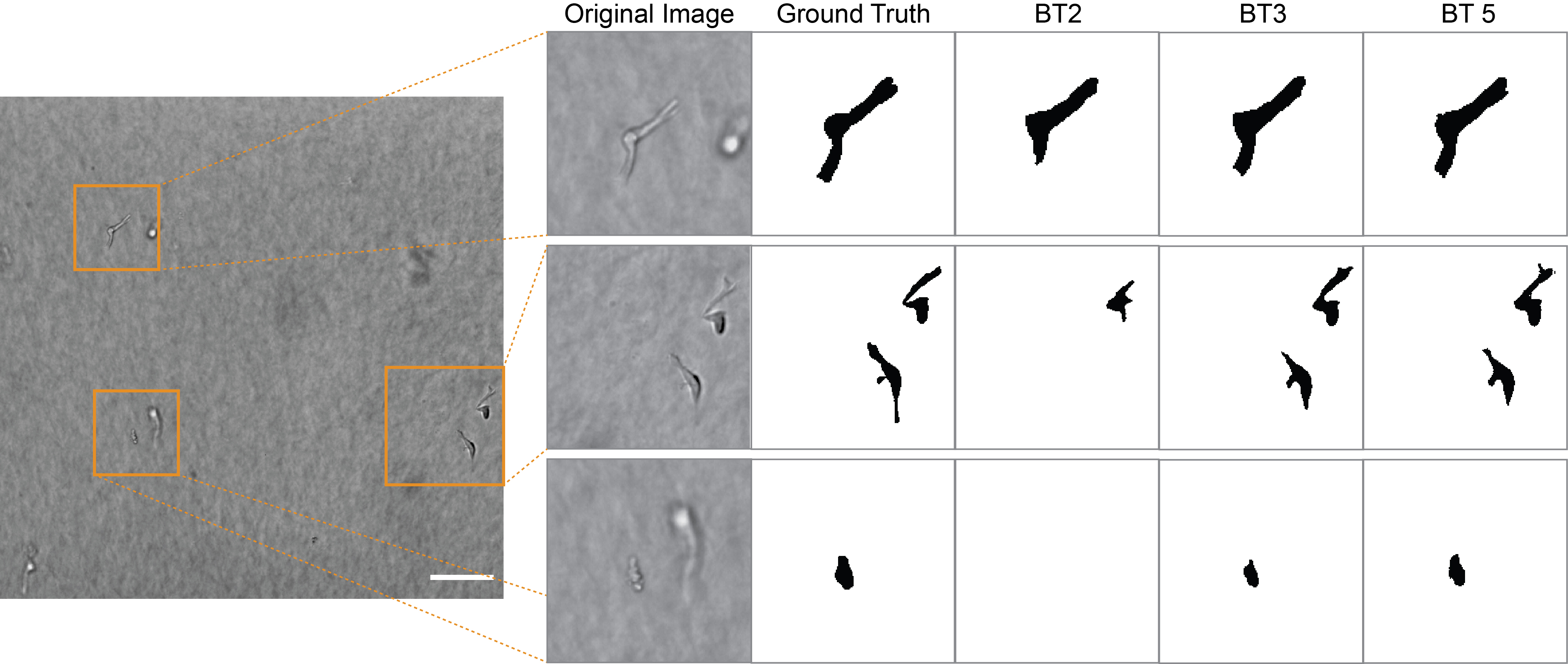}
    \caption{(Left) Phase-contrast microscopy image. Scale bar of $100\:\mu$; (Right) Zoomed crops of the region enclosed by the boxes in the full images; (From left to right) Original phase-contrast microscopy crop; Ground Truth (annotated in-focus cells are labeled in black); Output of the trained models B$2$, B$3$ and B$5$. }
    \label{fig: result bt2, bt3 and bt5}
\end{figure}
\begin{table}[h]
\begin{center}
\begin{tabular}{l*{7}{|c}}
        & Pools & Convolutional filters & Batch Size & Transfer learning & Fine tuning \\
\hline
 B$1$ & $5$ & $16-32-64-128-256$ & $1$ &  $0.473$ & $0.430$ \\
 B$2$ & $4$ & $25-50-100-200$ & $1$ &  $0.411$ & $0.455$\\
 B$3$ & $4$ & $25-50-100-200$ & $2$ &  $0.550$ & $0.551$\\
 B$4$ & $4$ & $50-100-200-400$ & $1$ &  $0.481$ & $0.446$ \\
 B$5$ & $3$ & $16-32-64$ & $2$  &  $0.437$ & $0.561$ \\
\end{tabular}
\caption{Convolutional neural network architecture and SEG for transfer learning and fine tuning. Note that the size of the convolutional layers only applies for the decoder path of the encoder-decoder. See Figure~\ref{fig: CNN architecture} for details about the architecture. Transfer learning and fine tuning processes are programmed for $2K$ epochs with a constant learning rate of $0.005$ and $0.0001$, respectively.}
\label{table: B and BT architectures}
\end{center}
\end{table} 

Unsurprisingly, a batch size larger than one improves the learning process making it smoother and more accurate. While the loss function for the training and validation datasets differs in shape and magnitude, the accuracy measured by the Jaccard index (Equation~\ref{eq: jaccard}) remains similar for both datasets, which could be due to the overlap between the training and validation sets. The CNN set up for B$4$ is the least stable, related to the larger size of the convolutional layers chosen and the small batch size. Table~\ref{table: B and BT architectures} shows that B$3$ and B$5$ are the most accurate CNNs. 

\section{Protrusion tip quantification}

Similar to the work of~\cite{Castilla2019}, we use the cell's skeleton and its end-points to identify the tips of the protrusions. When annotated manually, biologists consider cell protrusions that are longer than $5\:\mu m$. We use the Geodesic distance transform to measure the distance between the cell centroid and each detected tip, and estimate the protrusion length.  Hence, spurious tips can be accordingly discarded using this longitudinal estimation. We set the minimum length to $20\:\mu m$. Note that the Geodesic distance transform includes the radius of the cell nuclei region as it is computed from the cell body centroid. Most cells show a roundish pattern in their nucleic region. As cells are either elongated or rounded, the nuclei area will always be the wider region of the cell. Hence, the centroid is determined by the location of the maximum value of the Euclidean distance transform (see Figures~\ref{fig: protrusion detection} and~\ref{fig: protrusion result}).

\section{Discussion and conclusions}

Training strategy plays a crucial role in DL model training with scarce annotated data. The proposed training data sampling is relatively easy to implement and resembles the approach described in~\cite{Ronneberger2015}. It can hardly worsen the final results and often achieves similar performances with simpler CNN architectures. We believe that most image processing DL workflows should integrate a similar arrangement. 

The results obtained when using pre-trained encoders are already quite promising in terms of true and false positive pixel classification, which suggests that using encoders pre-trained on biomedical images could accelerate and improve the learning process. However, the outcome of the fine-tuned models improved, especially for those training schedules with batch sizes larger than one (see Tables~\ref{table: B and BT architectures}). Unfortunately, the batch sizes we could use were limited by the resources to train the CNN (see the number of parameters for each configuration in the Appendix). We use depth-wise separable convolutions to increase the field of view of the network at a lower parameter cost. Nevertheless, the receptive field of a pixel in our networks varies from $194 \text{ pixels}\times194 \text{ pixels}$ to $230 \text{ pixels}\times230 \text{ pixels}$, depending on the encoder-decoder depth chosen. Hence, reducing the spatial input size to allow larger batch sizes can prevent the network from visualizing the entire cell body without being affected by the padding artifacts of the convolutional layers. 

Another critical point in the image processing design is the trade-off between image resolution and network configuration. The thinnest details, i.e., the cell protrusions, are the primary source of error for the segmentation. We attribute these errors to the poor image resolution to draw cell protrusions and their large variability. Such low resolution hinders the learning of essential features and voids the effect that those pixels may have in the loss function or gradients. A simple computational experiment to test it could be to upsample the size of the images and repeat the training process.  

Thanks to the binary cell segmentation, we obtain accurate results for the cell tracking in low-resolution $2$D phase-contrast microscopy images. Due to the low density of cells in our images, they could be tracked easily with the last version of TrackMate~\cite{tinevez2017trackmate, Ershov2021Trackmate} which already processes object instances. The combination of whole cell shape segmentation and tracking is already a promising image processing tool for studying new clinical strategies to mitigate the metastatic phenotypes. 

\subsection*{Acknowledgments}
This work was partially funded by Ministerio de Ciencia, Innovación y Universidades, Agencia Estatal de Investigación, under Grant PID2019-109820RB-I00, MCIN / AEI / 10.13039/501100011033/, co-financed by European Regional Development Fund (ERDF), "A way of making Europe" (AMB); BBVA Foundation under a 2017 Leonardo Grant for Researchers and Cultural Creators (AMB); the US National Institutes of Health under Grants UO1AG060903 (DW) and U54CA143868 (DW). We also want to acknowledge the support of NVIDIA Corporation with the donation of the Titan X (Pascal) GPU used for this research.


\clearpage

\appendix
\textbf{Appendix}
\section{Microscopy data specifications}

\begin{figure}[ht!]
\centering
\includegraphics[width=0.55\textwidth]{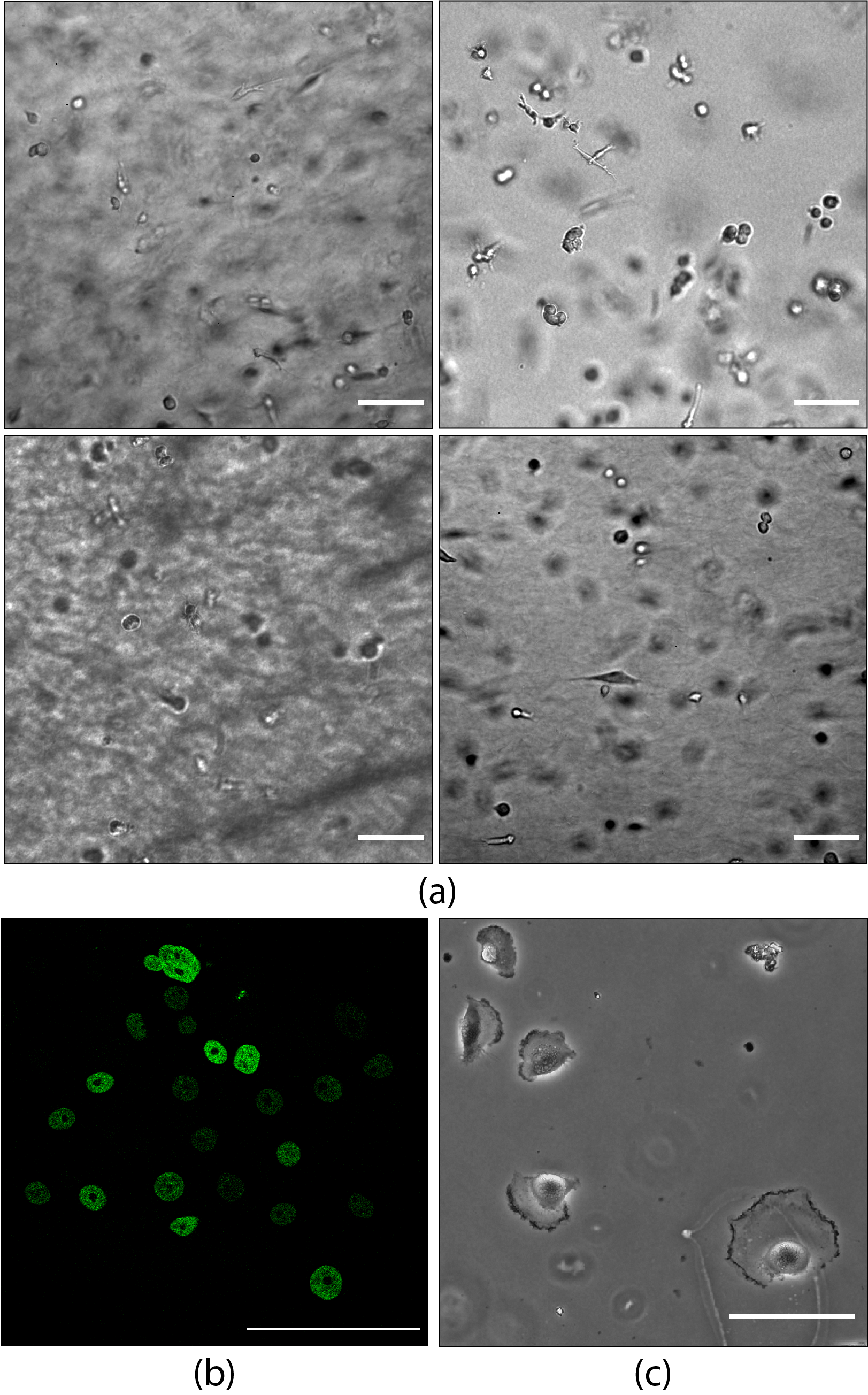}
\caption[Differences in microscopy image data acquisition and modality.]{ (a) Phase contrast microscopy images of cancer cells (MDA-MB-$231$) migrating in a $3$D collagen type I matrix. The collagen gels, cell culture and acquisition of each image was performed by a different researcher; (b) Fluorescence microscopy image of GFP-GOWT$1$ mouse stem cells migrating in 2D and fixed in paraformaldehyde~\cite{Ulman2017}; (c) Phase contrast microscopy image of glioblastoma-astrocytoma (U373) cells on a polyacrylamide substrate~\cite{Ulman2017}. Scale bars of $100\:\mu m$ in all the images.}
\label{fig: labexperiment differences}
\end{figure}

Human fibrosarcoma HT1080WT (ATCC) cells at low cell densities were embedded in $3$D collagen type I matrices ($5,000-10,000$ cells per $500\:\mu \mbox{mL}$ of collagen matrix) and placed on independent plates. They were imaged with a Cascade 1K CCD camera (Roper Scientific) mounted on a Nikon TE2000 microscope with a $10$X objective lens (i.e., low magnification). The time-lapse videos were recorded every $2$~minutes on a focal plane of at least $200\:\mu m$ away from the bottom of the culture plates to diminish edge effects. All the videos covered a field of view of $ 809\:\mu m \times 810\:\mu m$ and a total of $16.7$ hours ($ 1002\:\mbox{pixels} \times 1004\:\mbox{pixels} \times 500\:\mbox{frames}$). 

The videos suffer of frame-to-frame drift artifacts due to the microscope objective's drive when acquiring the temporal frames of each well in the plate. Because this feature is present in all the videos, we corrected the drift by applying an affine registration based on the compensation of the image correlation similarity frame-by-frame with the \texttt{StackReg} plugin (\url{http://bigwww.epfl.ch/thevenaz/stackreg/}) for ImageJ~\cite{Schneider2012, Schindelin2012, schroeder2020imagej, Rueden2017}. Then, the videos were cropped to curate border artifact, resulting in a final size of $983\:\mbox{pixels} \times 985\:\mbox{pixels} \times 500\:\mbox{pixels}$. See Figure \ref{fig: labexperiment differences} for sample images.

\section{Details for the cell segmentation deep learning model}

\subsection{Convolutional neural network blocks and layers}
\begin{figure}[htp!]
\centerline{\includegraphics[width=1\textwidth]{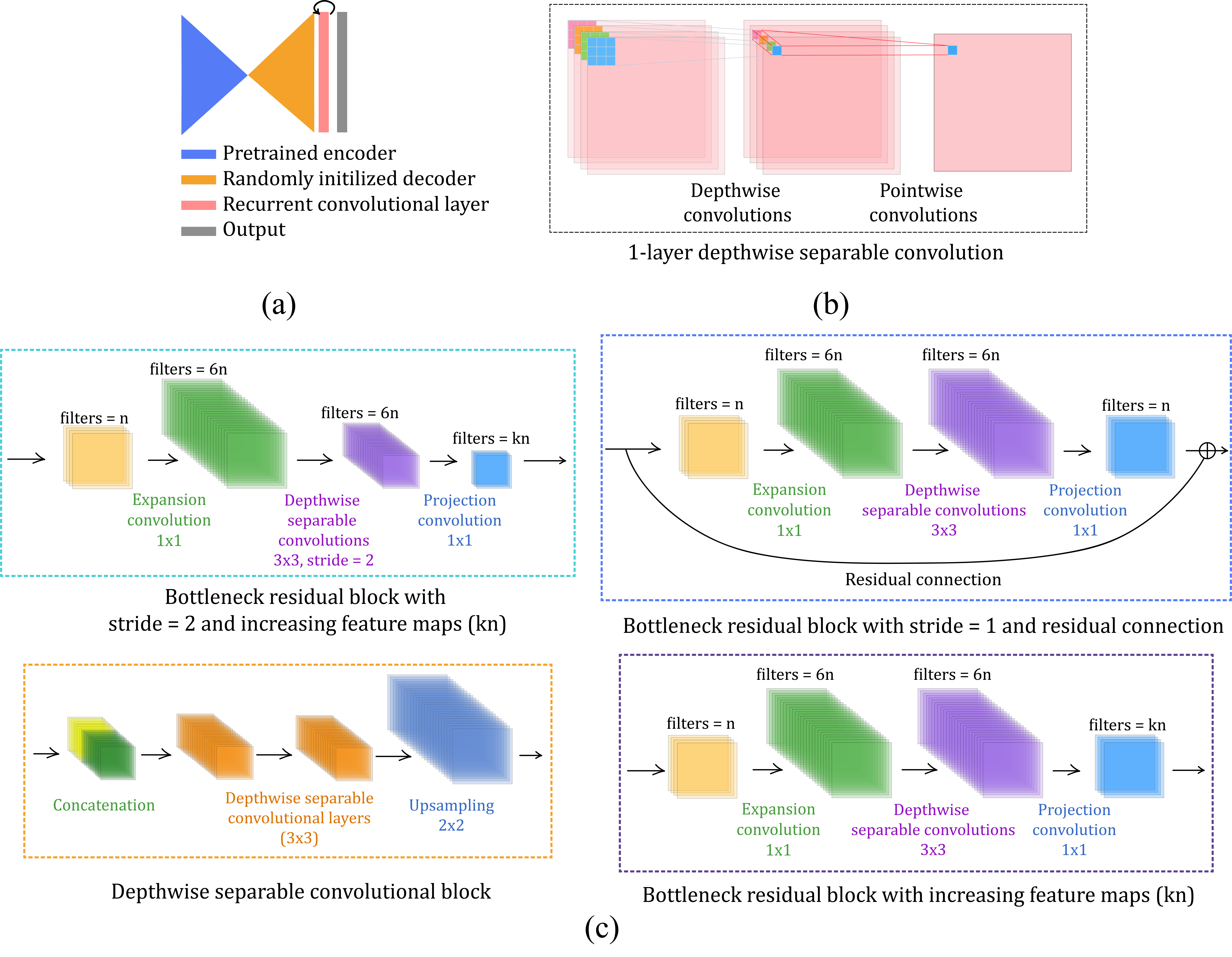}}
\caption[Recurrent convolutional neural network components.]{Recurrent convolutional neural network components: a) Single and multi-output modalities of the encoder-decoder described in Figure~\ref{fig: CNN architecture}. A MobileNetV2 trained on the ImageNet dataset~\cite{deng2009imagenet} was used as a pre-trained encoder. The decoder and the ConvLSTM units were randomly initialized; b) Description of a depth-wise separable convolution: (1) each channel of the input is filtered independently (depth-wise convolution), and (2) the resulting filtered channels are processed with a kernel of size $1\times 1 \times c$, $c$ being the number of channels in the input. If the depth-wise separable convolution has $n$ filters, the described process is repeated $n$ times; c) Description of each convolutional block used in the CNN architecture of Figure~\ref{fig: CNN architecture}. The expansion and the depth-wise separable convolutions are followed by batch normalization and a ReLU activation function in the encoder path. The projection convolutions are only followed by a batch normalization layer and are not activated. The depth-wise separable convolutions in the decoder are followed by batch normalization, an ELU activation function, and use dropout.}
\label{fig: CNN blocks}
\end{figure}

The main blocks and layer of the recurrent neural network component are give in Figure \ref{fig: CNN blocks}.

\subsection{Loss function and accuracy metrics}

The categorical cross-entropy loss function is given as:

\begin{equation}
    \mathcal{L}_{cell}(x,y) = - \sum\limits_{i=1}^{C=2}x_i\log(y_i),
    \label{eq: loss cell}
\end{equation}
where $C$ is total number of classes (background and foreground), $x_i$ is the class in the ground truth image and $y_i$ the score given by the CNN.

Adaptive moment estimation (ADAM) is defined as follows:
\begin{equation}
    m_t = \beta_1 m_{t-1} + (1-\beta_1)g_t, v_t = \beta_2 m_{t-1} + (1-\beta_2)g^2_t, 
    \label{eq: adam}
\end{equation}
where $m_t$ and $v_t$ are the estimates of the first moment and the second moment of the gradients, respectively, and $g_t$ refers to the gradient value at the time point $t$. We chose the default values given in Keras: $\beta_1 = 0.9$ and $\beta_2 = 0.999$.

The Jaccard index of the foreground is given as follows:
\begin{equation}
 JC = \dfrac{|X \cap Y | }{|X \cup Y |}
 \label{eq: jaccard}
\end{equation}
where $X$ is the ground truth and $Y$ is the output segmentation.

The Evaluation Software provided in the Cell Tracking Challenge~\cite{Ulman2017}\footnote{\url{http://celltrackingchallenge.net/evaluation-methodology/}} calculates segmentation (SEG) and tracking (TRA) accuracy measures. SEG is computed as the Jaccard index between the labeled objects in the ground truth and the resulting masks when the former covers more than $50\%$ of the output mask. Otherwise (less than $50\%$ of overlap), it sets the segmentation measure to zero. This accuracy measure is called SEG. As the length of the test videos is different, we averaged the SEG values of all the videos, Equation~\ref{eq: ctc jaccard}

\begin{equation}
   \overline{SEG}_{test} = \dfrac{1}{\|N\|}\sum\limits_{i=1}^{N} SEG_{i}(X_i, Y_i)
    \label{eq: ctc jaccard}
\end{equation}
where $X_i$ and $Y_i$ is video $i$ in the ground truth and the network's output, respectively, and $N$ is the total number of videos in the test set. TRA relies on acyclic oriented graphs matching (AOGM)~\cite{Matula2015}, calculated as follows:

\begin{equation}
    \mathrm{TRA} = 1- \frac{\min(\mathrm{AOGM_D}, \mathrm{AOGM_0})}{\mathrm{AOGM_0}}
\end{equation}
where AOGM$_{D}$ is the cost of transforming a set of nodes provided by the algorithm into the set of Ground Truth (GT) nodes, and AOGM$_0$ is the cost of creating the set of GT nodes from scratch (i.e., it is AOGM$_D$ for empty results). TRA behaves as an accuracy measure with values normalized to the $[0,1]$ interval. The final tracking measure TRA is the average of the TRA values obtained for each video

\begin{equation}
 \overline{TRA}_{test} = \dfrac{1}{\|N\|}\sum\limits_{i=1}^{N} TRA_{i}(X_i, Y_i).
  \label{eq: ctc tra}
\end{equation}

\subsection{Number of parameters of the convolutional neural networks}

See Table~$2$. 

\begin{table}[hp!]
\begin{center}
\begin{tabular}{c|c|c|c|}
        & Transfer learning & Fine tuning & Total \\
\hline
 B$1$ &  $227,266$ & $477,250$ & $490,530$  \\
 B$2$ &  $188,358$ & $280,326$ & $288,370$  \\
 B$3$ &  $188,358$ & $280,326$ &  $288,370$ \\
 B$4$ &  $622,008$ & $713,976$ & $723,520$  \\
 B$5$ &  $83,430$ & $99,270$ &  $101,714$ \\
 \hline
\end{tabular}
 \label{table:numberofpar}
\caption{Number of trainable parameters for each convolutional neural network architecture during transfer learning and fine tuning. Total values include trainable and non trainable parameters (convolutions' biases).}
\end{center}
\end{table} 

\subsection{Effect of data sampling in model training}

\begin{figure}[htpb]

    \centering
    \includegraphics[width=0.7\textwidth]{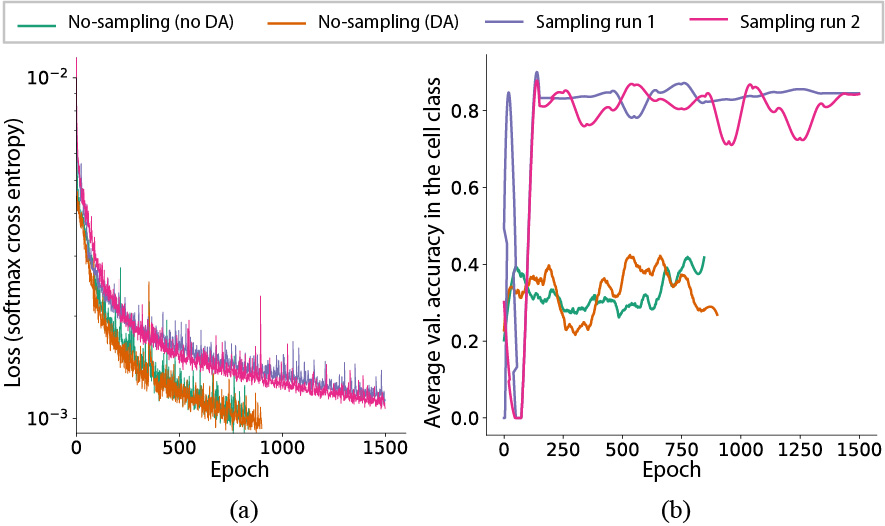}
    \caption[Effect of training data sampling strategy.]{Effect of training data sampling strategy. MobileUNet network~\cite{George2018, howard2017mobilenets} trained with the same parameters and changing the training data sampling strategy. (a) Softmax cross-entropy loss function for each training epoch. (b) Averaged accuracy of the foreground (pixels classified as cells) for the validation dataset on each $50$ epochs during the training. DA: data-augmentation.}
    \label{fig:training pdf}
\end{figure}

In Figure~\ref{fig:training pdf}, we show a sanity experiment performed with our sampling strategy. We trained the same network architecture with three different approaches: (1) On each iteration the data generator crops a random patch from an image and introduces it into the network (no-sampling (no DA)); (2) same as before but the patch is randomly transformed for data augmentation  (no-sampling (DA)); (3) first the original image is randomly transformed and then, a patch is cropped using a sampling probability distribution function over the pixels in the image. The loss function when there is no sampling probability distribution function (no-sampling) is lower than when it is applied (sampling run $1$ and sampling run $2$). However, the accuracy is less than $0.5$ for the foreground in those cases, indicating that the network classifies most of the pixels as background. During the training, CNN average the loss function values for all the pixels in the input patch. Because the foreground-background ratio is quite low, classifying all the pixels as background leads to a fickle local minimum. When using a sampling probability distribution function, such optimal modes are avoided. Moreover, it takes much longer until the loss function converges to a value similar to the one for which all the pixels are classified as background. The result, thus, is that the network learns to classify those poorly represented pixels with very little programming and mathematical effort.

\subsection{Stability of the learning process}

\begin{figure}[htpb]
    \centering
    \includegraphics[width=1\textwidth]{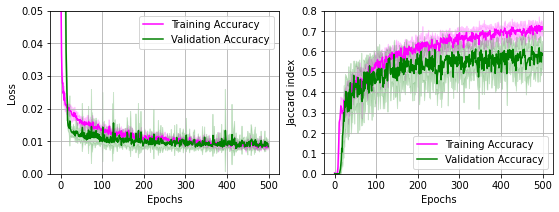}
    \caption{Replication of the learning process for the architecture B$5$ with exactly the same decoder initialization but randomly updating the training patches.}
    \label{fig:modelseed_1}
\end{figure}

\begin{figure}[htpb]
    \centering
    \includegraphics[width=1\textwidth]{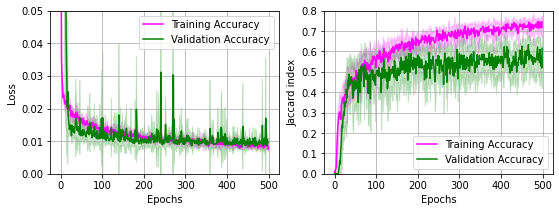}
    \caption{Replication of the learning process for the architecture B$5$ with a random decoder initialization but fixed training data patches.}
    \label{fig:modelseed_2}
\end{figure}

See Figures \ref{fig:modelseed_1} and \ref{fig:modelseed_2}. 

\subsection{Learning process of different model configurations}

\begin{figure}[hpb]
    \centering
    \includegraphics[width=1\textwidth]{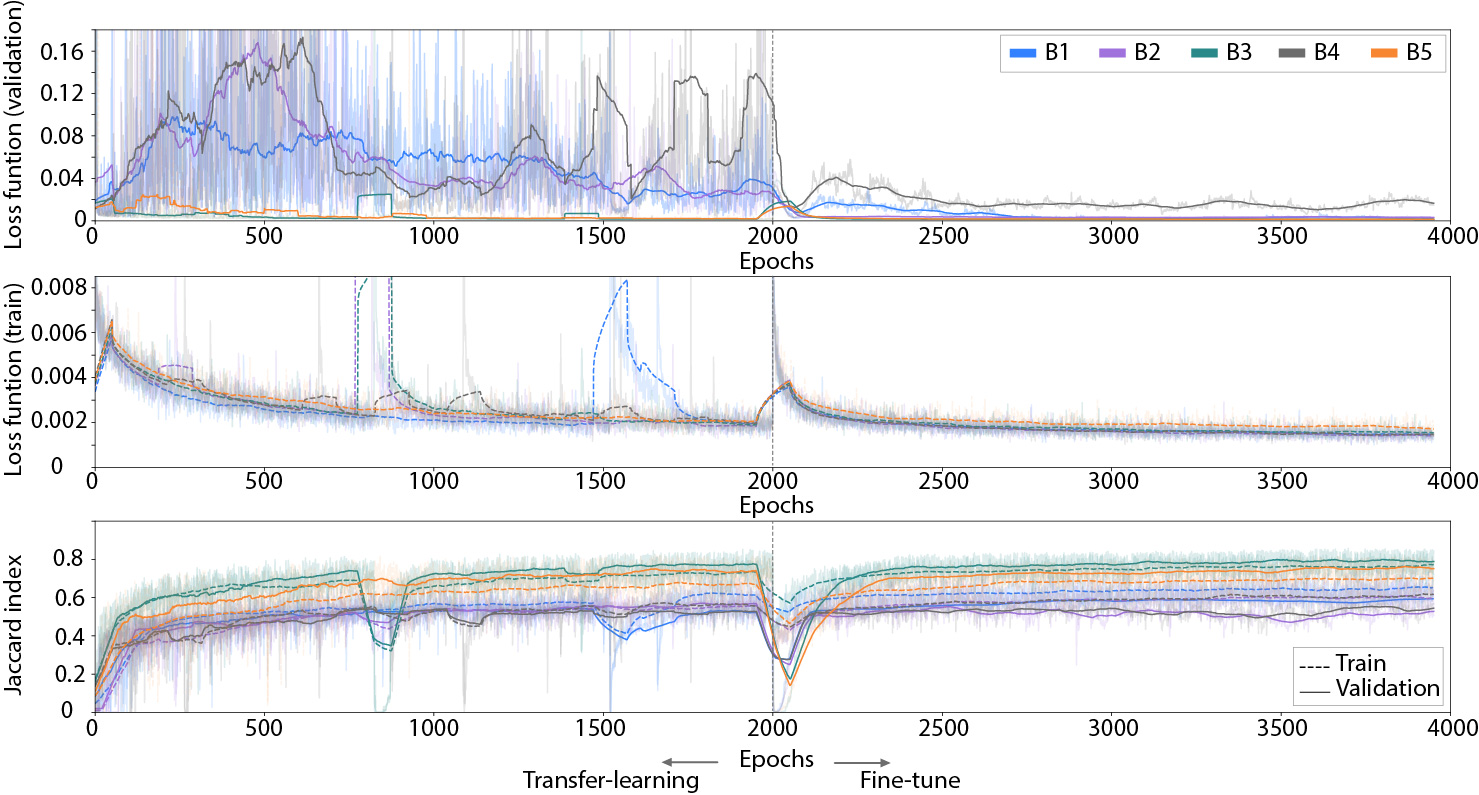}
    \caption[Learning process of the evaluated neural network architectures.]{Learning process of the evaluated neural network architectures: (a) Architectures trained to segment cells (B$1$, B$2$, B$3$, B$4$, and B$5$). During the first $2K$ epochs, the pre-trained encoder was frozen and the decoder is trained using a learning rate of  $0.005$. During the remaining $2K$ epochs, the gradients are back-propagated to all the weights in the model using a learning rate of $0.0001$. Training curves are smoothed to improve visualization.}
    \label{fig:modeltraining}
\end{figure}

See Figure \ref{fig:modeltraining}.

\section{Protrusion tip quantification}

\begin{figure}[htpb]
    \centering
    \includegraphics[width=0.75\textwidth]{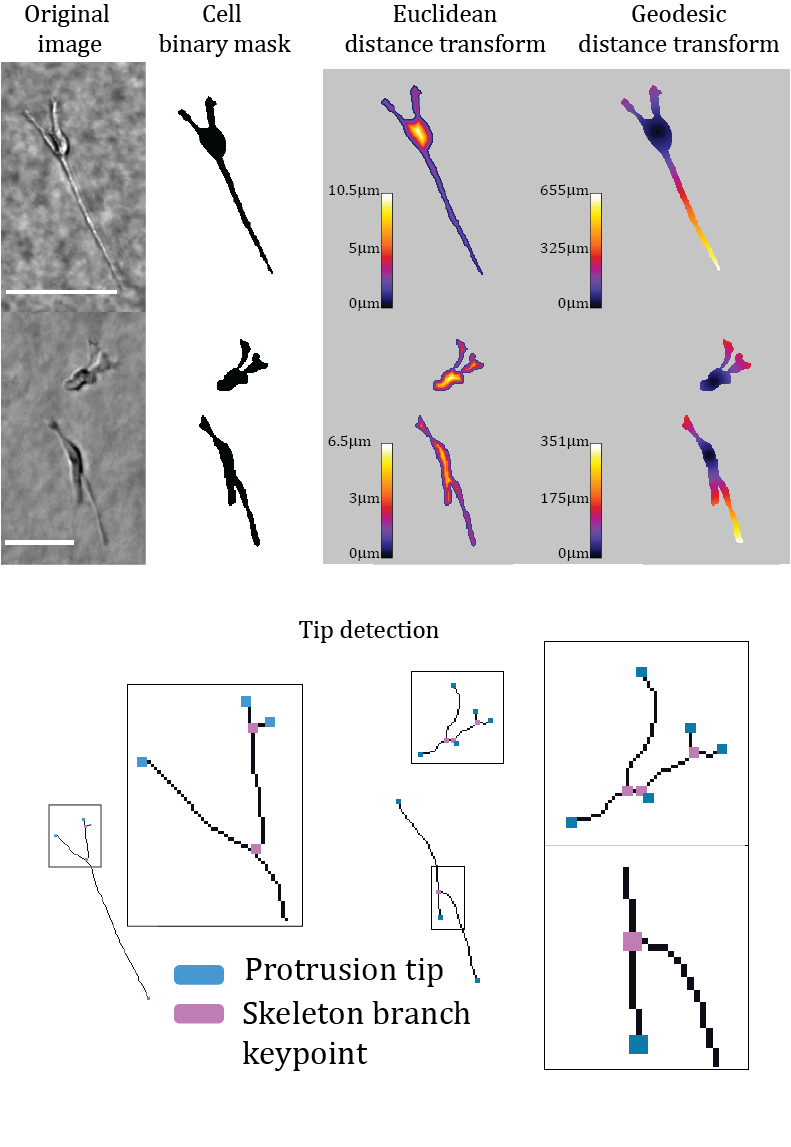}
    \caption{(Top) Image processing schema based on the cellular shape for the detection of protrusion tips. (Bottom) Sample images for the detection of protrusion tips and the skeleton branch keypoints. The images on the right correspond to the zoomed crops of the regions enclosed in the boxes on the left. Scale bars of $100\:\mu m$ and $50\:\mu m$ for the first and the second row, respectively.}
    \label{fig: protrusion detection}
\end{figure}

\begin{figure}[htpb]
    \centering
    \includegraphics[width=0.3\textwidth]{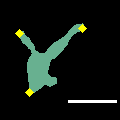}
    \includegraphics[width=0.3\textwidth]{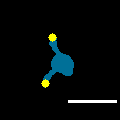}
    \includegraphics[width=0.3\textwidth]{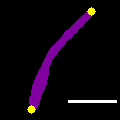}
    \caption{Result of the protrusion tip detection for different cell morphologies. The tips are labelled in yellow and the cell body in green, blue and violet, respectively. Scale bars of $40\:\mu m$.}
    \label{fig: protrusion result}
\end{figure}

See Figures \ref{fig: protrusion detection} and \ref{fig: protrusion result}.

\end{document}